# Continuous Function Structured in Multilayer Perceptron for Global Optimization

Heeyuen Koh†*

†Soft Foundry Institute, Seoul National University, Gwanak-ro 1, Gwanak-gu,
Seoul 08826, Republic of Korea

Corresponding Author *hy_koh@snu.ac.kr

**Abstract** The gradient information of multilayer perceptron with a linear neuron is modified with functional derivative for the global minimum search benchmarking problems. From this approach, we show that the landscape of the gradient derived from given continuous function using functional derivative can be the MLP-like form with ax+b neurons. In this extent, the suggested algorithm improves the availability of the optimization process to deal all the parameters in the problem set simultaneously. The functionality of this method could be improved through intentionally designed convex function with Kullack-Liebler divergence applied to cost value as well.

## I. INTRODUCTION

Multilayer perceptron(MLP) can afford various functionality like the process of human cognition or XOR logic gate which is out of the scope for continuous function. As the versatility of neural network structured in MLP has been extended to solve the system dealing with physics and chemistry whose matter of priorities is in the precision and high likelihood of the results, elucidating the functional space in MLP structure could be an emerging issue in the physics related research field. Accumulating the information into neural network form from partial differential equations or governing equation of the system which is once expressed as continuous function needs to be explicitly proved from the capability of the artificial neuron in multilayer perceptron.

Currently, neurons in MLP have shown its functionality for better classification ability by Fan et al.[1] which shows improved quality of classification using quadratic neuron in MLP. Functional space that neural network system can reach seems to be severely dependent on the unit of the MLP, the neuron as well as how it is built. Likewise, quantitative and mathematically organized protocol to build MLP to learn the functionality from continuous functions would accelerate the artificial intelligence to be more reliable tool in scientific research field.

The aspect that we can explore in this paper is the similarity or analogy of hierarchical structure of continuous function to MLP using functional derivative. As mentioned by Bengio[2], the cost function which is evaluating the MLP with its loss is a functional rather than a function. The functional whose variable is a function, has been one of major description in modern physics as well as its derivative as introduced by Parr and Yang.[3] Unlike the derivative of a function to its variable, the functional derivate as shown in Eq. (1) is integral with a function $\phi$ which indicate the direction of derivative in function space.

$$\lim_{\epsilon \to 0} \left[ \frac{F(f+\epsilon\phi)-F(f)}{\epsilon} \right] = \left( \frac{d}{d\epsilon} F(f+\epsilon\phi) \right)_{\epsilon \to 0},$$
$$= \int \frac{\delta F}{\delta f} \phi(x) dx. \quad (1)$$

$F$ is a functional, $f$ is continuous function defined on $x \in R$. In this paper, an algorithm based on the functional derivative is adapted to resolve the parameter optimization in back propagation instead of chain rule. Couple of global benchmark problems are shown to be solved.

The hierarchical structure of a continuous function $\rho_N$ can be written with $\rho_i(x; \theta_i)$ with the parameter set $\theta_i = \{\theta_{ij}\}, j = 0, \ldots, N_{\theta_i}$ as following:

$$\rho_N = \rho_{N-1}\left(\rho_{N-2}\left(\rho_{N-3}(\ldots \rho_1(x; \theta_1))\right)\right). \quad (2)$$

$\theta_i$ can be the set of all the parameter used in $\rho_i$ and $x \in R_N$. $N_{\theta_i}$ is the number of parameters that $\rho_i$ has. In this section, the functional derivative to optimize either of parameter set $\theta_i$ or the variable $x$ in $\rho_N$ is introduced in three subsections. First, how functional derivative can manage the gradient from the $\rho_N$ is explained with the given convex function in subsection A. In following subsection B, the component of the gradient from functional derivative is introduced and shown why the structure of gradient from each layer is identical to MLP. Lastly, three benchmarking problems from global minimum search are introduced to validate the suggested methods in subsection C.

## II. METHODS

### A. Functional derivative for cost functional

Cost function $X_{cost} = f(\rho_N - \rho_0)$ is the functional for evaluating the difference between $\rho_N$ and $\rho_0$.[2] When the nonlinearity in neural network from its cost function to offer maximum likelihood estimation from probability theory shapes nonconvex landscape, it is single parameter optimization which is adapted for continuous function of layers with hierarchy as Eq. (1). The landscape of each parameter in $\rho_i$ can be varied due to cross entropy decided by other parameter in its optimization process so that the optimization based on its landscape needs heuristic mind.

In this paper, we tried to restrain the result of the next step of optimization to be reside on the landscape given from the convex function that we designed using functional derivative. First, we designate that the cost function of $X_{cost}$ is drawn in any convex function. Then the cost functional defined for $\rho_N$ is as below:

$$F = \int X_{cost}(\rho_N) d\rho_N. \qquad (3)$$

As defined function $X_{cost}(\rho_N)$ as the convex function whose gradient has the information towards minimum $\rho_{cost} = \rho_N - \rho_0 = 0$, the amount of differentials towards minimum is given by functional derivation in Eq. (1) as below:

$$\delta F_{cost} = \int \frac{\delta F}{\delta f} \phi(x) d\rho_N, \qquad (4)$$

When cost function is a square function as $X_{cost} = \rho_{cost}^2$ where $\rho_{cost} = \rho_N - \rho_0$ and designate $f$ as $\rho_N$, the term $\frac{\delta F}{\delta f}$ in Eq. (1) becomes:

$$\frac{\delta F}{\delta f} = \frac{\delta \int X_{cost} d\rho_N}{\delta \rho_N} = \rho_{cost}, \qquad (5)$$

therefore, the quantification of functional derivative of $X_{cost}$ towards the minimum point of square along the function $\rho_{cost}$ is as below:

$$\delta F_{cost} = \int \frac{\delta F}{\delta f} \phi(\rho_N) d\rho_N = 2\int \rho_{cost}^2 d\rho_N \qquad (6)$$

, where $\phi(x) = \rho_{cost}$, and the result of Eq. (5) is $\delta F_{cost} = C\rho_{cost}^3$ where C is a constant and should be 2/3. The same result of Eq. (6) can be deduced from $f$ as $\rho_N^2$ and $\phi(x) = \rho_{cost}^2$ in which case the functional derivative is following the convex curve.

As conventionally used cost function, $\phi(x)$ can be equal to any order of $\rho_N$ since it is an arbitralily given. This indicate the functional derivative can be along any function defined by the combination of polynomials as well.

In reverse manner, sigmoid function can be the result of functional derivative $\delta F$ from certain form of convex curve. This convex from sigmoid or other activation function could be deduced a convex curve using simple ODE. The full derivation is in Appendix including the curvature of convex induced by sigmoid. This indicates that the hidden layers with activation function is, on the other aspect from functional derivative, can generate a fitting to certain value given from upper layer to be minimized. This will be discussed further in Section IV. Further process to specify the update for parameter optimization is in next subsection.

### B. Gradient in MLP with ax+b neuron

#### 1) Functional derivative for each layer

From Eq. (3) ~ (6), we could access the amount of $\rho_N$ to be fitted for optimization on the given convex. This implicate the value from $\rho_N$ is proportionally linked to the result of cost function on the convex curve. The same approach between $\rho_N$ and $\rho_i$ which is the functional derivative as below:

$$\delta F_i = \delta \rho_N / \delta \rho_i \phi(x) d\rho_i, \qquad (7)$$

where $i = 1, \ldots, N-1$ offers the information the proportionality between the value of cost function and the layers $\rho_i$, which are composing the $\rho_N$. An arbitrary function $\phi(x)$ can be $\rho_i$ or the function of $\rho_i$ as introduced in the previous section.

Linking $\delta F_i$ and $\delta F_{cost}$ in Eq. (5) finishes the proportionality between $X_{cost}$ to $\rho_i$. $\delta F_i$ and $\delta F_{cost}$ can be merged into $\delta F = \delta F_{cost} \delta F_i$ which has the information on how much $\delta \rho_i$ is changed from proportionality between $\rho_i$ and $\rho_{cost}$. $F$ offers the information about the distribution of cost function $X_{cost}$ on how much $X_{cost}$ can be varying on the convex from $\rho_i$. Then, the question on how to incorporate the information on $\delta F$ to the parameter set or the variable to conduct the optimization process is remained.

Since the approach for each layer of $\rho_i$ is different from Taylor expansion, the updating the parameter can be compensated by including the chain rule term since Eq. (1) does not provide the same information.

*2) Functional derivative for parameter optimization*

For the function, $\rho_i(x; \boldsymbol{\theta}_i)$, with variable $x$ and parameter set $\boldsymbol{\theta}_i$, there could be two different approaches to adapt the information from $\delta F$.

Firstly, there is a traditional way of using Taylor expansion from chain rule, $\delta \theta = \eta \frac{\partial X_{cost}(\rho_N)}{\partial \rho_i} \cdot \frac{\partial \rho_i}{\partial \theta}$ where $\eta$ is a learning rate. The blind point on this method is well-known local minima where one of the terms $\frac{\partial X_{cost}(\rho_N)}{\partial \rho_i}$ or $\frac{\partial \rho_i}{\partial \theta}$ is closed to zero so that $\delta\theta$ becomes zero before it reaches to global minimum.

On the contrary, the functional derivative of $\delta F = \delta F_{cost} \delta F_i$, the chain rule can be included from Taylor expansion or functional derivative according to the optimization target which should be either of $\boldsymbol{\theta}_{i\_op} = \{\theta_{ij}\}, j = 0, \dots, N_{\theta_i}$ or $x_{op}$. Lowercase character $_{op}$ means they are the answer of optimization process for $\boldsymbol{\theta}_i$ or $x$. Applying Taylor expansion for the update from $\delta F$ is simply multiplying $\delta F$ with $\partial \rho_i / \partial \theta_{ij}$ or $\partial \rho_i / \partial x$. In case of functional derivative, $\delta \rho_i / \delta \theta$ or $\delta \rho_i / \delta x$ is applied to $\delta F = \delta F_{cost} \delta F_i$ so that the $\delta F_{total}$ becomes as below:

$$\delta F_{total} = \delta F_{cost} \delta F_i \int \frac{\delta \rho_i}{\delta \theta_{ij}} \theta_{ij} dx, \forall x \quad (8)$$

$\delta F_{total} \neq 0$ assures the proportionality between $\theta_{ij}$ or $x$ to $X_{cost}$ in case of optimizing the minimum for $\theta_{ij}$ or $x$, respectively. Functional derivative to find $x_{op}$ where $X_{cost}$ has its minimum makes Eq. (6) to be $\delta F_{total} = \delta F_{cost} \delta F_i \int \frac{\delta \rho_i}{\delta x} x dx, \forall x$. Functional derivative of Eq. (6) for N-1 layers in total offers information on how much $\delta F_N$ can be adjusted from each layer, therefore, the landscape of the gradient from each layer offers more information than $\partial X_{cost} / \partial \theta$.

Former has the form directly proportional to its own value so that it can be regarded as $a\theta_{ij}$ as $aX + b$ neuron. In this case, the higher order of $\theta_{ij}$ suggested in Eq. (6) is possible as well, however it is disregarded in the trial. The latter from Taylor expansion become bias in the neuron which is $b$ in $aX + b$. The $\delta F_i$ can be used without the integration but multiplied with $\partial \rho_i / \partial \theta$ or just the result from the chain rule of $F_N$ by $\theta_{ij}$ can be another source of bias as well. This notation or analogy is from the concept of generalization using functional derivative where $\rho_i$ with $\theta$. These approaches are applicable on any continuous function parameter optimization when $\delta F / \delta \rho$ exists. The authors remain a notation that the hierarchical structure of $aX + b$ neuron is what multilayered perceptron has.

*C. Application to global minimum search problems*

So far we introduced the derivation to turn the gradient information from functional derivative of continuous function into multilayer perceptron ($MLP_f$). It is applied to two global minimum benchmarking problem known as Cross Table Leg(CTL) and De Ville Grasser 02 (DVG). The problem definitions for CTL and DVG are in Appendix. They are global minimum benchmarking problems which has not been solved by Particle Swarm Optimization up to the most recent reference[4,5]. As shown in Fig. 1(a) CTL function has very thin curve except $x_j = 0$, i=1 or 2, yet as each layer of CTL is managed into $ax + b$ form, global minimum search is successfully completed for several random initial points. DVG in Fig. 2(b) which is also known for its difficulty to be optimized is tried with $MLP_f$ and conventional Taylor expansion separately. The results for both CTL and DVG are in Table 1.

The functional derivative from the last layer $\rho_N - \rho_0$, is finalized as cubic of cost which has similar curve to activation function. The sigmoid is derived as functional derivative from a convex function as shown in the appendix. The difference of this cost function derived as the activation-like curvature is confirmed using LJ potential energy cluster optimization. 13 particles interacting with LJ potential energy $E_{LJ} = -4\epsilon \left( \left(\frac{\sigma}{r}\right)^6 - \left(\frac{\sigma}{r}\right)^{12} \right)$ is supposed to be optimized in icosahedron. The $E_{LJ}$ is decomposed into $MLP_f$ with two different convex function. One is from a square function and the other curvature is derived sigmoid function.

Lastly, the cost between $\rho_N$ and $\rho_0$ is evaluated using Kulleck-Liebler Divergence which is definition as $\rho_{cost_{KDL}} = \log(\rho_{cost} + off) - \log(\rho_0 + off)$. Off is any positive value to make sure the inside of log to be >0. The difference of adding KDL as the cost value is introduced in the results section.

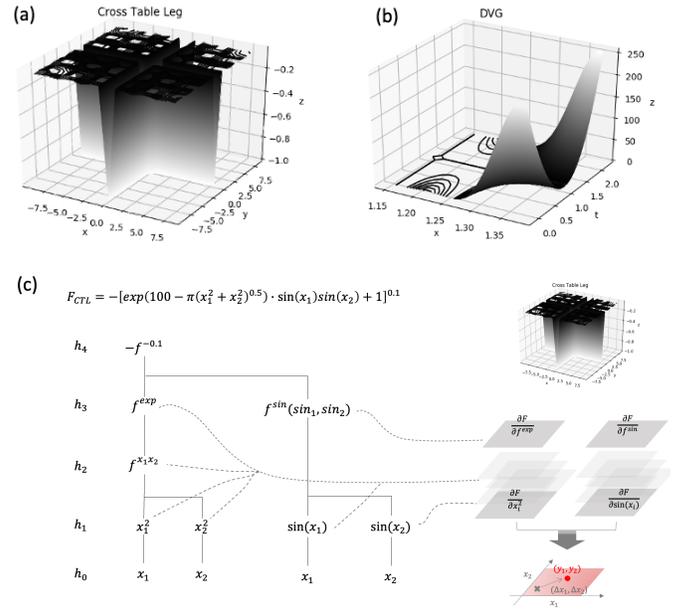

Fig. 1. (a) Cross Table Leg function in two dimensions, (b) De Viller Grasser 02 function, (c) the structure of CTL to be prepared as $MLP_f$. Each layer $h_i$ in hierarchical structure of CTL is decomposed into gradient using functional derivative (gray shade) and merged into the information for $x_1$ and $x_2$.(red shade)

## III. RESULTS

As mentioned in II-C, we tried to optimize three global minimum search problem. Each of them is CTL, DVG and LJ cluster optimization problem.

### A. CTL and DVG

As shown in the Table 1. The optimization results from various initial points have well fitted into the global minimum. In case of DVG, we found a point which has very slow optimization, otherwise all the cases are well proved to be optimized. Even though the optimization is very slow, it never remained at certain point which means it is not in local minimum. For the comparison, we added Taylor expansion case. As expected, the optimization process is less effective compared to MLP$_f$ and more points are remained to be trapped in local minimum. The code including the derived MLP$_f$ from each continuous function is uploaded in https://github.com/ieebon/GM_BP.

Table 1. The result of optimization for CLT and DVG function.

| Function | Initial | Final | GM | # of steps |
|---|---|---|---|---|
| Cross Leg Table | (-7,5) | (1e-5,1e-5) | (0,0) | < 1e5 |
|  | (7,5) |  |  |  |
|  | (7,-5) |  |  |  |
|  | (-7,-5) |  |  |  |
| De Ville Grasser 02 | [50,50,50,50,1] | [52.99, 1.28, 5.21, 31.42, 0.51] | [53, 1.27,3.01,2.13 ,0.57] | > 1e6 |
|  | [10,10,10,10,10] | [51.96, 1.29, 5.97, -0.55, 0.51] |  |  |
|  | [10,10,10,10,10]* | [53.96, 1.26, 2.8, 2.32, 0.51] |  |  |
|  | [0.2,0.2,0.2,0.2,0.2] | [53.81, 1.27, 3.01, 2.13, 0.51] |  |  |

### B. LJ potential energy cluster

The arrangement of the particles interacting with LJ potential energy has several local minima. The initial coordinate of 13 particles is intentionally given from one of local minimum conformation and observed its progress with Gradient descentdent method (GDM) from Taylor expansion and MLP$_f$ separately.

As shown in Fig. 2, GDM from Taylor expansion has no optimization from its local minimum. MLP$_f$ with cost function defined in square($f_{x2}$) and the convex derived from sigmoid($f_{sig}$) are well optimized the icosahedron with 13 particles, however the potential energy of fully structured icosahedron cluster is varied from the minimum energy which is -44.32 eV. This difference is analyzed in next section. Note that $f_{sig}$ completes the process much faster than $f_{x2}$.

Both convex functional needs KDL otherwise the optimization did not operate properly in case of $f_{sig}$ as shown in Fig. 2 (b). In case of $f_{x2}$ without KDL, still the icosahedron structure is optimized, but its energy could not reach close to the global minimum level. The initial and final coordinate of icosahedron is in Fig. 2a and the formulation and code are supplementary data and https://github.com/ieebon/GM_BP, respectively.

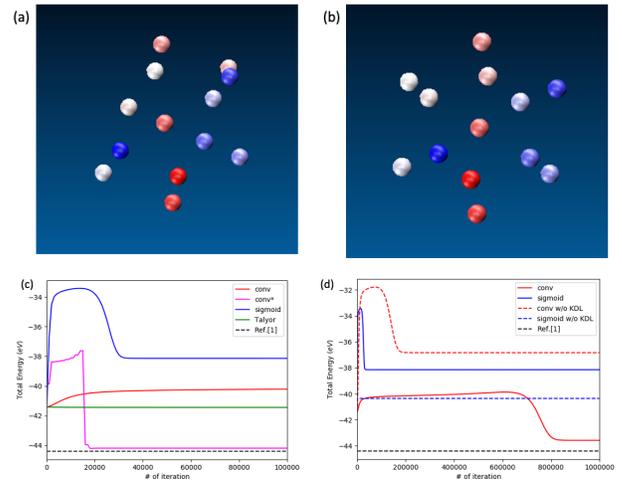

Fig. 2. 13 particle icosahedron optimization using MPL$_f$ for Lennard-Johnson potential energy function. (a) initial configuration of 13 particles in local minimum which has no further optimization using Taylor expansion, (b) The result from MPL$_f$ constructed using functional derivative. The code is included in https://github.com/ieebon/GM_BP (c) The potential energy calculated during optimization process using different cost functions which are square function (red and magenta), sigmoid function (blue) and Taylor expansion with square function (green). The Reference [6] shows the icosahedron at this configuration has -44.32 eV (dashed, black). The result of conv* in magenta represent the fitting from multiple terms which is including functional derivative for the terms in factorization. Only square cost function is closed to fully optimized condition, yet the speed of sigmoid function in processing icosahedron structure is rapid so that it completes the conformation in 1e4 iteration. (d) The trial without KDL is shown with dashed line. The level of potential energy optimized without KDL has not reached to right level.

## IV. DISCUSSION

$aX + b$ form in MLP$_f$ is investigated for the gradient landscape of continuous function. The gradient landscape of any continuous function can be expressed as elaborated model of MLP$_f$. It has certain similarity to Universal approximation theorem [7] since the continuous function has diffeomorphism with its gradients space composed with functional derivative. In reverse manner, we could conjecture that the complete function which has its gradient landscape with MLP$_f$ should be re-constructed with $aX^2 + bX + c$ as Fan et al., has suggested in classification using multilayer perceptron.[1]

For technically adapting the MLP$_f$ into global optimization benchmarking problems, few necessary conditions have been revealed. The convex function which is given at the end of the MLP$_f$ is transformed into cubic or sigmoid function with KDL as cost value. In this way, the optimization process could be activated with its efficiency and precision. Compared to Taylor expansion, the availability of MLP$_f$ as global optimization is clearly shown through 3 different optimization problems. When the function is factorized, each term in the multiplication offers different gradient, including all the possible terms has enhanced the precision in LJ cluster optimization case.

From the functional derivative on arbitrary convex function at the end of MLP$_f$, we could confirm that the sigmoid function is the results of a convex curve and its functional differential. This indicate the sigmoid or any rectifiable function can be the process of optimization to minimize the given value during back propagation as driving the result of the neuron near zero. How the rectifiable functions in hidden layers like CNN can boost the functionality of deep learning in various learning process is beyond the scope of this paper.

## V. CONCLUSION

In this paper, we have shown that the functional derivative derived from a given arbitrary convex function and the hierarchy of the continuous function forms MLP structure with ax+b neuron. The functional derivative from convex function is the cubic or sigmoid function of the cost variable adjusted the gradient from MLP$_f$ to be fitted on the convex.

Including KDL as a cost value three global minimum benchmarking problems and LJ energy cluster optimization have been solved by MLP$_f$. All the benchmarking problem which has been known for its difficulties to optimize using PSO of Taylor series has well resolved from MLP$_f$.

## APPENDIX A.

When the convex in functional derivative like Eq. (5) offers a sigmoid function, that convex should satisfy following condition:

$$\int \frac{\partial X_{cost}}{\partial f} \phi(x) \, dx = \frac{1}{1+\exp(-\rho_{cost})}, \quad (A1)$$

$$f = X_{cost}, \quad (A2)$$

$$\phi(x) = f_{conv}, \quad (A3)$$

$$\frac{\partial X_{cost}}{\partial f} \phi(x) = \phi(x) = \frac{\partial}{\partial x}\left(\frac{1}{1+\exp(-\rho_{cost})}\right), \quad (A4)$$

$$\phi(x) = \frac{\exp(-\rho_{cost})}{(1+\exp(-\rho_{cost}))^2} \frac{\partial \rho_{cost}}{\partial x}. \quad (A5)$$

$\phi(x)$ is defined on $\rho_N$ and $\rho_{cost}$ is the norm between $\rho_N$ and $\rho_{op}$ which forms convex curve. For optimization,

$\rho_{cost} = \|\rho_{op} - \rho_N\|$ should be adapted to make convex of $\phi(x)$.

## APPENDIX B.

Function of CTL [4,5] :
$$F_{CTL} = -[\exp(100 - \pi(x_1^2 + x_2^2)^{0.5}) \cdot \sin(x_1)\sin(x_2) + 1]^{0.1} \quad (8)$$

Function of DVG [4,5]:
$$f_{DVG} = \sum \left[ x_1 x_2^{t_i} \tanh(x_3 t_i + \sin(x_4 t_i))\cos(t_i e^{x_5}) - y_i \right]^2 \quad (9)$$

, where $t = 0.1 \, i$ and $i = 1,..,24$ with

$y_i = 53.81(1.27_i^t \tanh(3.01 t_i + \sin(2.13 t_i)) \cos(t_i e^{0.507}))$

## ACKNOWLEDGMENT

This research is funded by Basic Science Research Program through the National Research Foundation of Korea(NRF) funded by the Ministry of Education (NRF-2020R1I1A1A01071567) and funded by Ministry of Science and ICT (NRF-2016M3D1A1021141).

*Clusters Containing up to 110 Atoms*", in *J. Phys. Chem. A 1997, 101, 5111-5116*.

[7] K. Hornick et al. "Multilayer Feedforward Networks are Universal Approximators", Neural Networks, Vol. 2, pp. 35Y-366, 1989.